\documentclass{article} 
\usepackage{iclr2020_conference,times}

\usepackage{amsmath,amsfonts,bm}

\def\eqref#1{equation~\ref{#1}}

\def\1{\bm{1}}

\def\vtheta{{\bm{\theta}}}

\def\vz{{\bm{z}}}

\DeclareMathAlphabet{\mathsfit}{\encodingdefault}{\sfdefault}{m}{sl}
\SetMathAlphabet{\mathsfit}{bold}{\encodingdefault}{\sfdefault}{bx}{n}

\usepackage{enumitem}
\usepackage{graphicx}
\usepackage{hyperref}
\usepackage{url}
\usepackage{booktabs}

\title{Towards Understanding Normalization in Neural ODEs}

\author{Julia Gusak, Larisa Markeeva, Talgat Daulbaev, Alexandr Katrutsa,\\ \textbf{Andrzej Cichocki \& Ivan Oseledets}\\
Skolkovo Institute of Science and Technology\\
Moscow, Russia\\
\texttt{\{y.gusak,l.markeeva,talgat.daulbaev\}@skoltech.ru}\\
\texttt{aleksandr.katrutsa@phystech.edu}\\
\texttt{\{a.cichocki,i.oseledets\}@skoltech.ru}
}

\iclrfinalcopy 
\begin{document}

\maketitle

\begin{abstract}
Normalization is an important and vastly investigated technique in deep learning. However, its role for Ordinary Differential Equation based networks (neural ODEs) is still poorly understood.
This paper investigates how different normalization techniques affect the performance of neural ODEs. 
Particularly, we show that it is possible to achieve $93\%$ accuracy in the CIFAR-10 classification task, and to the best of our knowledge, this is the highest reported accuracy among neural ODEs tested on this problem.
\end{abstract}

\section{Introduction}
Neural Ordinary Differential equations (neural ODEs) are proposed in~(\cite{chen2018neural}) and model the evolution of hidden representation with ordinary differential equation (ODE).
The right-hand side of this ODE is represented with some neural network.
If one considers classical Euler scheme to integrate this ODE, then ResNet-like architecture~(\cite{he2016deep}) will be obtained.
Thus, Neural ODEs are continuous analogue of ResNets.
One of the motivation to introduce such models was assumption on smooth evolution of the hidden representation that can be violated with ResNet architecture.
Also, in contrast to ResNet models, Neural ODEs share parameters of the ODE right-hand side between steps to integrate this ODE.
Thus, Neural ODEs are more memory efficient. 


Different normalization techniques were proposed to improve the quality of deep neural networks.  
Batch normalization~(\cite{ioffe2015batch}) is a useful technique when training a deep neural network model. 
However, it requires computing and storing moving statistics for each time point. 
It becomes problematic when a number of time steps required for different inputs vary as in recurrent neural networks~(\cite{hochreiter1997long,cooijmans2016recurrent,ba2016layer}), or the time is continuous as in neural ODEs.
We apply different normalization techniques (\cite{salimans2016weight, miyato2018spectral,ba2016layer}) to Neural ODE models and report results for the CIFAR-10 classification task.
The considered normalization approaches are compared in terms of test accuracy and ability to generalize if a more powerful ODE solver is used in the inference. 

\section{Background}

The main ingredient of the neural ODE architecture is the ODE block.
The forward pass through the ODE block is equivalent to the solve the following initial value problem (IVP)
\begin{equation}
\begin{cases}
\dfrac{\mathrm{d} \vz}{\mathrm{d} t} = f(\vz(t), t, \vtheta), \quad t \in [t_0, t_1] \\ 
\vz(t_0) = \vz_0,  
\label{eq:fwd_1}
\end{cases}
\end{equation}
where $\vz_0$ denotes the input features, which are considered as initial value.
To solve IVP, we numerically integrate system~(\ref{eq:fwd_1}) using ODE solver. Depending on the solver type different number of RHS evaluations of (1) are performed.
Initial value problem~(\ref{eq:fwd_1}) replaces Euler discretization for the same right-hand side that arises in ResNet-like architectures.
One part of the standard ResNet-like architecture is the so-called ResNet block, which consists of convolutions, batch normalizations, and ReLU layers.
In practice, batch normalization is often used to regularize model, make it more robust to training hyperparameters and reduce internal covariate shift~(\cite{shimodaira2000improving}).
Also, it is shown that batch normalization yields smoother loss surface and makes neural network training faster and more robust~(\cite{santurkar2018does}).
In the context of neural ODEs training, previous studies applied layer normalization~(\cite{chen2018neural}) and batch normalization~(\cite{gholami2019anode}) but did not investigate the influence of these layers on the model performance.
In this study, we focus on the role of normalization techniques in neural ODEs.
We assume that proper normalization applied to the layers in ODE blocks leads to the higher test accuracy and smoother dynamic.

According to~\cite{luo2018differentiable}, different problems and neural network architectures require different types of normalization. 
In our empirical study, we investigate the following normalization techniques to solve the image classification problem with neural ODE models.
\begin{itemize}[leftmargin=*]
\item \emph{Batch normalization} (BN;~\cite{ioffe2015batch}) is the most popular choice for the image classification problem, we discuss its benefits  in the above paragraph.
\item \emph{Layer normalization}~(LN; \cite{ba2016layer}) and \emph{weight normalization}~(WN; \cite{salimans2016weight}) were introduced for RNNs.
We consider these normalizations as a ppropriate candidates for incorporating in neural ODEs since they showed its effectiveness for RNNs that also exploit the idea of weights sharing through time. 
\item \emph{Spectral normalization}~(SN; \cite{miyato2018spectral}) was proposed for generative adversarial networks. 
It is natural to consider SN for neural ODEs since if the Jacobian norm is bounded by~$1$, one may expect better properties of the gradient propagation in the backward pass. 
\item We also trained neural ODEs without any normalization (NF).
\end{itemize}

To perform back-propagation, we use ANODE~(\cite{gholami2019anode}) approach. 
This is a memory-efficient procedure to compute gradients in neural ODEs with several ODE blocks.
This method exploits checkpointing technique at the cost of extra computations.

\section{Numerical Experiments}
This section presents numerical results of applying different normalization techniques to neural ODEs in the CIFAR-10 classification task.
Firstly, we compare test accuracy for neural ODE based models with different types of normalizations. 
Secondly, we present an \emph{$(\mathcal{S}, n)$-criterion} to  estimate quality of the trained neural ODE-like model.
The source code is available at GitHub repository\footnote{\url{https://github.com/juliagusak/neural-ode-norm}}.

In our experiments we consider neural ODE based models, which are build by stacking standard layers and ODE blocks. 
After replacing ResNet block with ODE block in ResNet4 model, we get
\begin{center}
conv $\to$ \emph{norm} $\to$ activation $\to$ ODE block $\to$ avgpool $\to$ fc 
\end{center}
an architecture, which we call \emph{ODENet4}. 
For this model we test different normalization techniques for \emph{norm} layer and inside the ODE block.
Similarly, by replacing in ResNet10 architecture ResNet blocks, which do not change spacial size, with ODE blocks, we get the following model: 
\begin{center}
    conv $\to$ \emph{norm} $\to$ activation $\to$ ResNet block $\to$ ODE block $\to$ ResNet block $\to$ ODE block $\to$ avgpool $\to$ fc,
\end{center}
which we call \emph{ODENet10}. 
In contract to ODENet4, this model admits different normalizations in place of the \emph{norm} layer, inside ResNet blocks and ODE blocks.


We use ANODE to train  considered models since it is more robust than the adjoint method (more details see in~\cite{gholami2019anode}).
In both forward and backward passes through ODE blocks we solve corresponding ODEs using Euler scheme. 
For the training schedule, we follow the settings from ANODE~(\cite{gholami2019anode}). 
In contract to ANODEDEV2~(\cite{zhang2019anodev2}), we include activations and normalization layers to the model.
We train considered models for 350 epochs with an initial learning rate equal to 0.1. 
The learning rate is multiplied by $0.1$ at epoch 150 and 300. 
Data augmentation is implemented. 
The batch size used for training is 512. 
For all experiments with different normalization techniques, we use the same settings.

\subsection{Accuracy}
In our experiments, we assume that normalizations for all ResNet blocks are the same, as well as for all ODE blocks.  Along with these two normalizations, we vary a normalization technique after the first convolutional layer.
We report test accuracy for different normalization schedules for ODENet10.
Table~\ref{tab:resnet10_acc} presents test accuracy given by ODENet10 model.
The best model achieves 93\% accuracy.
It uses batch normalization after the first convolutional layer and in the ResNet blocks, and layer normalization in the ODE blocks.
Also, we observe that the elimination of batch normalization after the first convolutional layer and from the ResNet blocks leads to decreasing accuracy to 91.2\%.
Such quality is even worse than the quality obtained with the model without any normalizations (92\%).

\begin{table}[!h]
    \centering
    \begin{tabular}{c|ccccc}
       ODE blocks  & BN & WN & SN & NF & LN \\
       \hline
       Accuracy@1  & $0.762$ & $0.925$ & $0.926$ & $0.927$  & $\bm{0.930}$
    \end{tabular}
    \caption{
    Comparison of normalization techniques for ODENet10 architecture on CIFAR-10.
    BN~-- batch normalization, LN~-- layer normalization, WN~-- weight normalization, SN~-- spectral normalization, NF~-- the absence of any normalization.
    To perform back-propagation, we exploit ANODE with a non-adaptive ODE solver. 
    Namely, we use Euler scheme with $N_t = 8$, where $N_t$ is a number of time steps used to solve IVP~(\ref{eq:fwd_1}). 
    The first row corresponds to the normalization in the ODE blocks.
    We use BN after the first convolutional layer and inside ResNet blocks, respectively.
    Standard ResNet10 architecture (only ResNet blocks are used) gives $\bm{0.931}$ test accuracy.
    }
    \label{tab:resnet10_acc}
\end{table}

\subsection{$(\mathcal{S}, n)$-criterion of dynamics smoothness in the trained model}

Since in neural ODEs like models, we train not only parameters of standard layers, but also parameters in the right-hand side of the system~(\ref{eq:fwd_1}), the test accuracy is not the only important measure.
Another significant criterion is the smoothness of the hidden representation dynamic that is controlled by the trained parameters of the right-hand side~(\ref{eq:fwd_1}).

To implicitly estimate this smoothness, we propose an \emph{($\mathcal{S}, n$)-criterion} that indicates whether more powerful solver induces performance improvement of the trained neural ODE model during evaluation.
Here, $\mathcal{S}$ denotes a solver name (Euler, RK2, RK4, etc) and $n$ denotes a number of the right-hand side evaluations necessary for integration of system~(\ref{eq:fwd_1}), which corresponds to the forward pass through the ODE block.
By more powerful solver we mean ODE solver that requires more right-hand side evaluations to solve~(\ref{eq:fwd_1}) than ODE solver used in training for the same purpose.
For example, assume one trains the model with Euler scheme and $n = 2$.
Then, we say that ODE block in trained model corresponds to smooth denamics if using Euler scheme with $n > 2$ during evaluation yields higher accuracy. 
Otherwise, we say that $(\mathcal{S}, n)$-criterion shows the absence of learned smooth dynamics.
Worth noting that the $(\mathcal{S}, n)$-criterion has limitation.
Namely, it requires the solution of IVP~(\ref{eq:fwd_1}) to be a Lipchitz function of the right-hand side~(\ref{eq:fwd_1}) parameters and inputs~(\cite{coddington1955theory}). 
Otherwise, we can not rely on this criterion since the closeness in the right-hand side parameters does not induce the closeness of features that are inputs to the next layers of the model.

In our experiments we consider ODENet4 architecture with four different settings of the Euler scheme: $n =  2, 8, 16, 32$.  
For each setting we have trained 10 types of architectures that differ from each other by the type of normalization we apply to the first convolutionl layer and convolutional layers in the ODE block. 
For example, the model named ``ODENet4 BN-LN (Euler, 2)'' means the following: we have used ODENet4 architecture, where after conv layer follows a BN layer, after each convolutional layer in the right-hand side~(\ref{eq:fwd_1}) follows an LN layer, and Euler scheme with 2 steps is used to propagate through the ODE block. 

For a fixed model trained with (Euler, $n_0$) solver we check the fulfillment of $(\mathcal{S}, n)$-criterion by evaluating its accuracy with more powerful solver.
In this case, we consider the following more powerful solvers: (Euler, $n$), (RK2, $n$) and (RK4, $n$), where $n > n_0$.

\begin{figure}[!h]
\centering
\includegraphics[width=\linewidth]{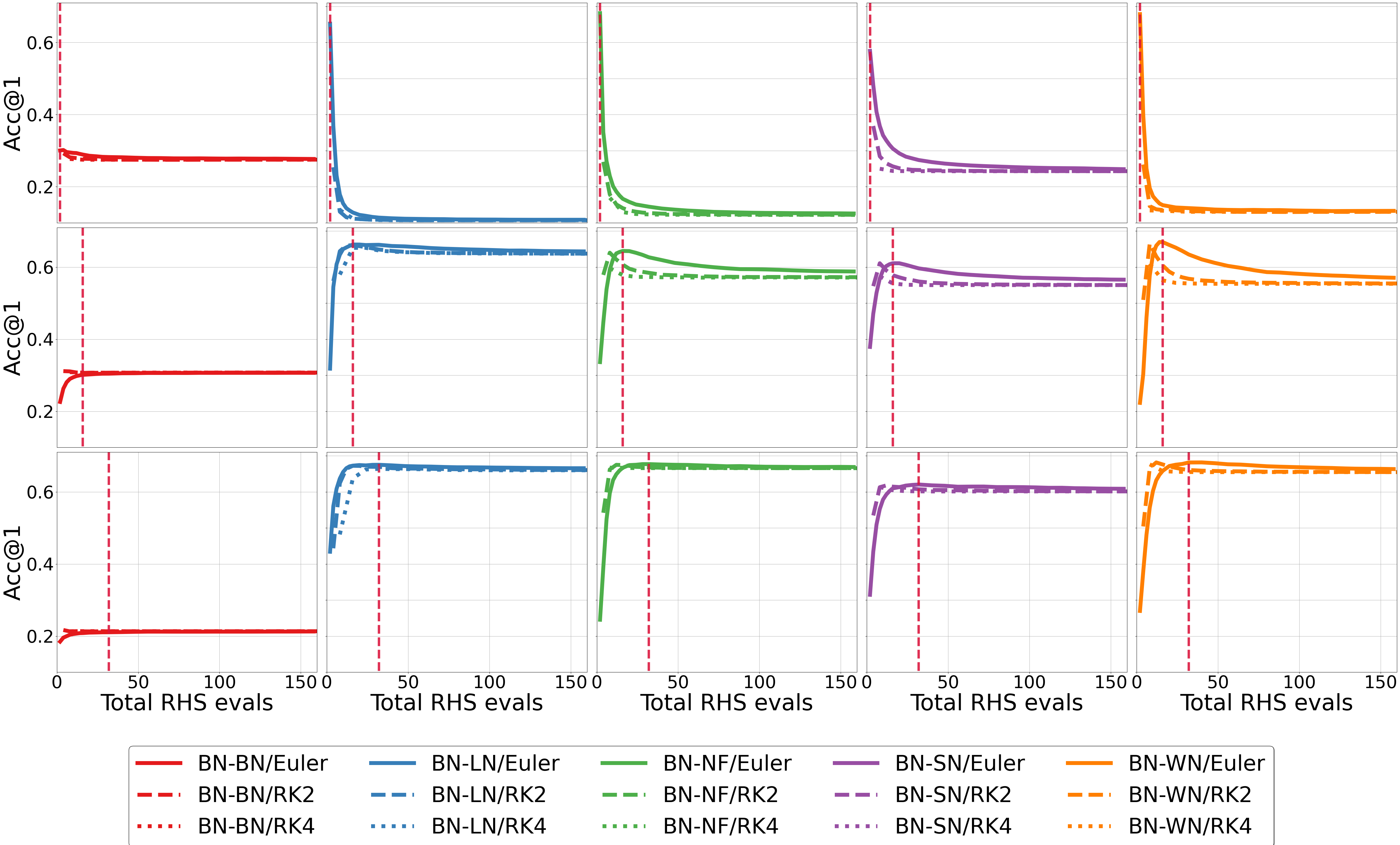}
\caption{Illustration of how the choice of ODE solver and normalizations during training implicitly affects the smoothness of learned dynamics. 
Each subplot corresponds to the model trained with a fixed ODE solver and normalization scheme. 
Models within one row have the same type of training solver ((Euler, $n$), $n = 2, 16, 32$ from top to bottom). 
Models within one column have the same normalization technique. 
For example, subplot in the third row and the  second column corresponds to the ODENet4 model trained with (Euler, 32) solver with BN after the first convolutional layer and LN after convolutional layers inside ODE block.
Lines of different style corresponds to different types of test solvers.
If model accuracy does not drop when the more powerful ODE solver is used, we conclude that, according to ($\mathcal{S}, n$)-criterion, the model provides a smooth dynamics.
For example, the model (Euler, 32) BN-LN trains a smooth dynamics, while (Euler, 2) BN-LN fails to do that.
Also, we can observe that to learn the smooth dynamics during training, for some normalization schemes less powerful solvers  are required.
If we compare BN-LN and BN-WN models, we can see that the first one learns smooth dynamics when Euler with $n = 16$ is used, but the latter one does that only for $n = 32$.
}
\label{fig:resnet4_acc}
\end{figure}

In Figure~\ref{fig:resnet4_acc}, we show how test accuracy given by ODENet4 model with different normalizations changes with varying ODE solvers to integrate IVP~(\ref{eq:fwd_1}) in ODE blocks.
Different line types correspond to different solver type (Euler, RK2, RK4), $x$-axis depicts the number of the right-hand side evaluations, while $y$-axis stands for the test accuracy.
These models were trained with Euler scheme and after that we use Euler, RK2 and RK4 schemes to compute test accuracy. 
Every row from top to bottom corresponds to $n=2,16,32$ used in Euler scheme. 

\section{Discussion and Further research}
We have empirically investigated the effect of normalization techniques to ODE based models. 
For different models, we have compared test accuracy as well as the ability to learn parameters that yield smooth dynamics of hidden representations. 
We have observed that both normalization and type of training solver affect the performance of the final model. 
Worth noting that pre-trained models, which are close in terms of test accuracy, can significantly differ when it comes to the smoothness of the hidden representations dynamics. 
In our further research, we plan to investigate how different tasks benefit from the presence of smoothness in hidden representations. 
Also, we will work on a more rigorous theoretical criterion that can be used to compare ODE based models,  considering  both neural networks and ODEs metrics.



\section*{Acknowledgement}
Sections 2 and 3 were supported by Ministry of Education and Science of the Russian Federation grant 14.756.31.0001.
High-performance computations presented in the paper were carried out on Skoltech HPC cluster Zhores.

\bibliography{iclr2020_conference}
\bibliographystyle{iclr2020_conference}


\end{document}